\documentclass[10pt,twocolumn,letterpaper]{article}

\usepackage{iccv}
\usepackage{times}
\usepackage{graphicx,subcaption} \graphicspath{ {Images/} }
\usepackage{bm,amsmath,amssymb,amsthm}
\usepackage{microtype}
\usepackage{url}
\usepackage{booktabs} 
\usepackage[ruled]{algorithm2e}
\usepackage{authblk}

\usepackage[pagebackref=true,breaklinks=true,letterpaper=true,colorlinks,bookmarks=false]{hyperref}

\iccvfinalcopy 

\def\iccvPaperID{112}

\ificcvfinal\fi
\begin{document}

\title{End-to-End Deep Convolutional Active Contours for Image Segmentation}

\author{Ali Hatamizadeh}
\author{Debleena Sengupta}
\author{Demetri  Terzopoulos}
\affil{Computer Science Department \\ 
University of California, Los Angeles, CA, USA}
\maketitle

\begin{abstract}
The Active Contour Model (ACM) is a standard image analysis technique
whose numerous variants have attracted an enormous amount of research
attention across multiple fields. Incorrectly, however, the ACM's
differential-equation-based formulation and prototypical dependence on
user initialization have been regarded as being largely incompatible
with the recently popular deep learning approaches to image
segmentation. This paper introduces the first tight unification of
these two paradigms. In particular, we devise Deep Convolutional
Active Contours (DCAC), a truly end-to-end trainable image
segmentation framework comprising a Convolutional Neural Network (CNN)
and an ACM with learnable parameters. The ACM's Eulerian energy
functional includes per-pixel parameter maps predicted by the backbone
CNN, which also initializes the ACM. Importantly, both the CNN and ACM
components are fully implemented in TensorFlow, and the entire DCAC
architecture is end-to-end automatically differentiable and
backpropagation trainable without user intervention. As a challenging
test case, we tackle the problem of building instance segmentation in
aerial images and evaluate DCAC on two publicly available datasets,
Vaihingen and Bing Huts. Our reseults demonstrate that, for building
segmentation, the DCAC establishes a new state-of-the-art performance
by a wide margin. 
\end{abstract}

\section{Introduction}

\begin{figure}
\centering
\includegraphics[width=\linewidth]{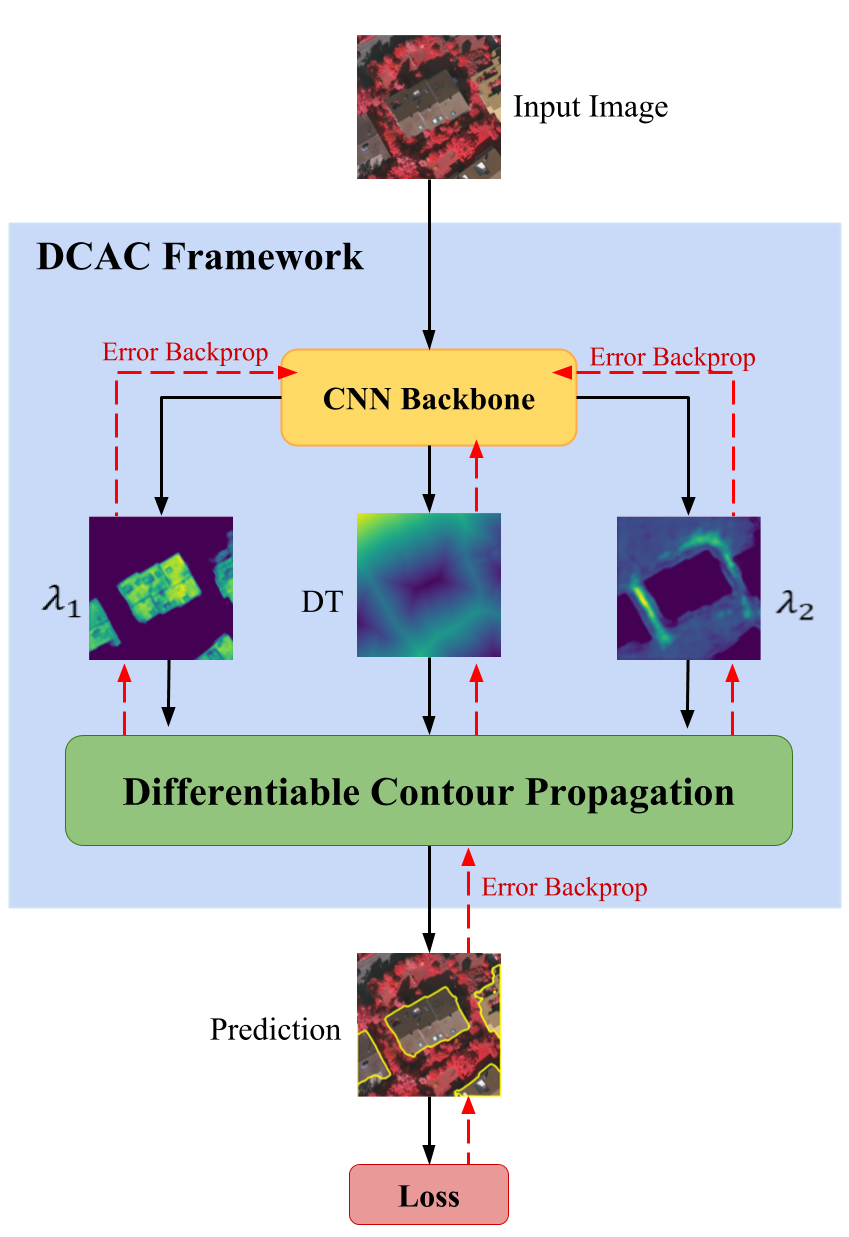}
\caption{DCAC is a framework the end-to-end training of an
automatically differentiable ACM and backbone CNN without user
intervention, implemented entirely in TensorFlow. The CNN learns to
properly initialize the ACM, via a generalized distance transform, as
well as the per-pixel parameter maps in the ACM's energy functional.}
\label{fig:framework}
\end{figure}
 
The ACM \cite{kass1988snakes} is one of the most influential computer
vision techniques. It has been successfully employed in various image
analysis tasks, including object segmentation and tracking. In most
ACM variants the deformable curve(s) of interest dynamically evolves
through an iterative procedure that minimizes a corresponding energy
functional. Since the ACM is a model-based formulation founded on
geometric and physical principles, the segmentation process relies
mainly on the content of the image itself, not on large annotated
image datasets, extensive computational resources, and hours or days
of training. However, the classic ACM relies on some degree of user
interaction to specify the initial contour and tune the parameters of
the energy functional, which undermines its applicability to the
automated analysis of large quantities of images.

In recent years, Deep Neural Networks (DNNs) have become popular in
many areas. In computer vision and medical image analysis, CNNs have been succesfully exploited for different segmentation tasks \cite{hatamizadeh2018automatic,hatamizadeh2019boundary,myronenko20193d}. Despite their tremendous success, the performance of
CNNs is still very dependent on their training datasets. In essence,
CNNs rely on a filter-based learning scheme in which the weights of
the network are usually tuned using a back-propagation error gradient
decent approach. Since CNN architectures often include millions of
trainable parameters, the training process relies on the sheer size of
the dataset. In addition, CNNs usually generalize poorly to images
that differ from those in the training datasets and they are
vulnerable to adversarial examples \cite{xie2017mitigating}. For image
segmentation, capturing the details of object boundaries and
delineating them remains a challenging task even for the most
promising of CNN architectures that have achieved state-of-the-art
performance on relevant bench-marked datasets
\cite{chen2017rethinking,he2017mask,zhao2017pyramid}. The recently
proposed Deeplabv3+ \cite{chen2018encoder} has mitigated this problem
to some extent by leveraging the power of dilated convolutions, but
such improvements were made possible by extensive pre-training and
vast computational resources---50 GPUs were reportedly used to train
this model.

In this paper, we aim to bridge the gap between CNNs and ACMs by
introducing a truly end-to-end framework. Our framework leverages an
automatically differentiable ACM with trainable parameters that allows
for back-propagation of gradients. This ACM can be trained along with
a backbone CNN from scratch and without any pre-training. Moreover,
our ACM utilizes a locally-penalized energy functional that is
directly predicted by its backbone CNN, in the form of 2D feature
maps, and it is initialized directly by the CNN. Thus, our work
alleviates one of the biggest obstacles to exploiting the power
ACMs---eliminating the need for any type of user supervision or
intervention.

As a challenging test case for our DCAC framework, we tackle the
problem of building instance segmentation in aerial images. Our DCAC
sets new state-of-the-art benchmarks on the \textit{Vaihingen} and
\textit{Bing Huts} datasets for building instance segmentation,
outperforming its closest competitor by a wide margin.

\section{Related Work}

\paragraph{Eulerian active contours:}
Eulerian active contours evolve the segmentation curve by dynamically
propagating an implicit function so as to minimizing its associated
energy functional \cite{osher2001level}. The most notable approaches
that utilize this formulation are the active contours without edges by
Chan and Vese \cite{chan2001active} and the geodesic active contours
by Caselles \textit{et al.} \cite{caselles1997geodesic}. The
Caselles-Kimmel-Sapiro model is mainly dependent on the location of
the level-set, whereas the Chan-Vese model mainly relies on the
content difference between the interior and exterior of the level-set.
In addition, the work by \cite{lankton2008localizing} proposes a
reformulation of the Chan-Vese model in which the energy functional
incorporates image properties in local regions around the level-set,
and it was shown to more accurately segment objects with heterogeneous
features.

\paragraph{``End-to-End'' CNNs with ACMs:}
Several efforts have attempted to integrate CNNs with ACMs
in an end-to-end manner as opposed to utilizing the ACM merely as a
post-processor of the CNN output. Le \textit{et al.}
\cite{le2018reformulating} implemented level-set ACMs as Recurrent
Neural Networks (RNNs) for the task of semantic segmentation of
natural images. There exists 3 key differences between our proposed
DCAC and this effort: (1) DCAC does not reformulate ACMs as RNNs and
as a result is more computationally efficient. (2) DCAC benefits from
a novel locally-penalized energy functional, whereas
\cite{le2018reformulating} has constant weighted parameters. (3) DCAC
has an entirely different pipeline---we employ a single CNN that is
trained from scratch along with the ACM, whereas
\cite{le2018reformulating} requires two \emph{pre-trained} CNN
backbones (one for object localization, the other for classification).
The dependence of \cite{le2018reformulating} on pre-trained CNNs has
limited its applicability. The other attempt, the DSAC model by Marcos
\textit{et al.} \cite{marcos2018learning}, is an integration of ACMs
with CNNs in a structured prediction framework for building instance
segmentation in aerial images. There are 3 key differences between
DCAC and this work: (1) \cite{marcos2018learning} heavily depends on
the \emph{manual initialization} of contours, whereas our DCAC is
fully automated and runs without any external supervision. (2) The ACM
used in \cite{marcos2018learning} has a parametric formulation that
can handle only a single building at a time, whereas our DCAC
leverages the Eulerian ACM which can naturally handle multiple
building instances simultaneously. (3) \cite{marcos2018learning}
requires the user to \emph {explicitly} calculate the gradients,
whereas our approach fully automates the direct back-propagation of
gradients through the entire DCAC framework due to its automatically
differetiable ACM.

\paragraph{Building instance segmentation:}
Modern CNN-based methods have been used with different approaches to
the problem of building segmentation. Some efforts have treated this
problem as a semantic segmentation problem
\cite{paisitkriangkrai2015effective,wang2016torontocity} and utilized
post-processing steps to extract the building boundaries. Other
efforts have utilized instance segmentation networks
\cite{Iglovikov_2018_CVPR_Workshops} to directly predict the location
of buildings.

\begin{figure*}
\centering
\subcaptionbox{Input
image}{\includegraphics[width=0.24\linewidth,height=0.24\linewidth]{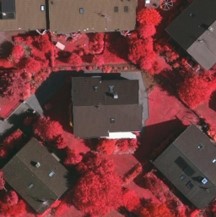}}\hfill
\subcaptionbox{Learned distance
transform}{\includegraphics[width=0.24\linewidth,height=0.24\linewidth]{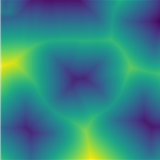}}\hfill
\subcaptionbox{$\lambda_1(x,y)$}{\includegraphics[width=0.24\linewidth,height=0.24\linewidth]{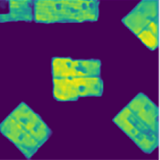}}\hfill
\subcaptionbox{$\lambda_2(x,y)$}{\includegraphics[width=0.24\linewidth,height=0.24\linewidth,trim={20
0 0 0},clip]{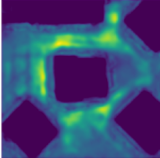}}
  \caption{Examples of learned distance transform, $\lambda_{1}$ and $\lambda_{2}$ maps for a given input image.}
  \label{fig:leanred_func}
\end{figure*}



\section{Level Set Active Contours}

First proposed by Osher and Sethian \cite{osher1988fronts} to evolve
wavefronts in CFD simulations, a level-set is an implicit
representation of a hypersurface that is dynamically evolved according
to the nonlinear Hamilton-Jacobi equation. In 2D, let $C(t) =
\big\{(x, y) | \phi(x, y,t) = 0 \big\}$ be a closed time-varying
contour represented in $\Omega\in R^2$ by the zero level set of the
signed distance map $\phi(x,y,t)$. Function $\phi(x,y,t)$ evolves
according to
\begin{equation}
  \begin{cases}
  \frac{\partial\phi}{\partial t}={|\nabla\phi|} \text{div}
  \Big(\frac{\nabla\phi}{|\nabla\phi|}\Big);  \\
  \phi(x,y,0)= \phi_{0}(x,y),     \\
  \end{cases}
\end{equation}
where $\phi(x,y,0)$ represents the initial level set.

We introduce a generalization of the level-set ACM proposed by Chan
and Vese \cite{chan2001active}. Their model assumes that the image of
interest $I(x,y)$ consists of two areas of distinct intensities. The
interior of $C$ is represented by the smoothed Heaviside function
\begin{equation}
\label{eq:heviside}
H_\epsilon(\phi)= \frac{1}{2} +
\frac{1}{\pi}\arctan\Big(\frac{\phi}{\epsilon}\Big)
\end{equation}
and $1-H_\epsilon$ represents its exterior. The derivative of
(\ref{eq:heviside}) is the smoothed Dirac delta function
\begin{equation}
\delta_\epsilon(\phi) = \frac{\partial H_\epsilon(\phi)}{\partial \phi}
= \frac{1}{\pi}\frac{\epsilon}{\epsilon^2+\phi^2}.
\end{equation}
The energy functional associated with $C$ is written as
\begin{equation}
\begin{split}
\label{eq:f_pc}
E(&\phi(x,y,t))= \\ & \int_\Omega \mu
\delta_\epsilon(\phi(x,y,t))|\nabla\phi(x,y,t)| + \nu
H_\epsilon(\phi(x,y,t)) \,dx\,dy \\ + & \int_\Omega
\lambda_1(x,y)(I(x,y)-m_1)^2H_\epsilon(\phi(x,y,t)) \,dx\,dy \\ + &
\int_\Omega \lambda_2(x,y)(I(x,y)-m_2)^2(1-H_\epsilon(\phi(x,y,t)))
\,dx\,dy,
\end{split}
\end{equation}
where $\mu$ penalizes the length of $C$ and $\nu$ penalizes its
enclosed area (we set $\mu=0.2$ and $\nu=0$), and where $m_1$ and
$m_2$ are the mean image intensities inside and outside $C$. We follow
Lankton \textit{et al.} \cite{lankton2008localizing} and define $m_1$
and $m_2$ as the mean image intensities inside and outside $C$ within
a local window around $C$.

Note that to afford greater control over $C$, we have generalized the
constants $\lambda_1$ and $\lambda_2$ used in \cite{chan2001active} to
parameter \emph{functions} $\lambda_1(x,y)$ and $\lambda_2(x,y)$ in
(\ref{eq:f_pc}). The contour expands or shrinks at a certain location
$(x,y)$ if $\lambda_2(x,y)>\lambda_1(x,y)$ or
$\lambda_2(x,y)<\lambda_1(x,y)$, respectively \cite{hatamizadeh2019deeplesion}. In DCAC, these
parameter functions are trainable and learned directly by the backbone
CNN. Fig.\ref{fig:leanred_func} illustrates an example of these
learned maps by the CNN.

Given an initial distance map $\phi(x,y,0)$ and parameter maps
$\lambda_1(x,y)$ and $\lambda_2(x,y)$, the ACM is evolved by
numerically time-integrating, within a narrow band around $C$ for
computational efficiency, the finite difference discretized
Euler-Lagrange PDE for $\phi(x,y,t)$; refer to \cite{chan2001active}
and \cite{lankton2008localizing} for the details.

\section{CNN Backbone}

\begin{figure*}
\includegraphics[width=\linewidth]{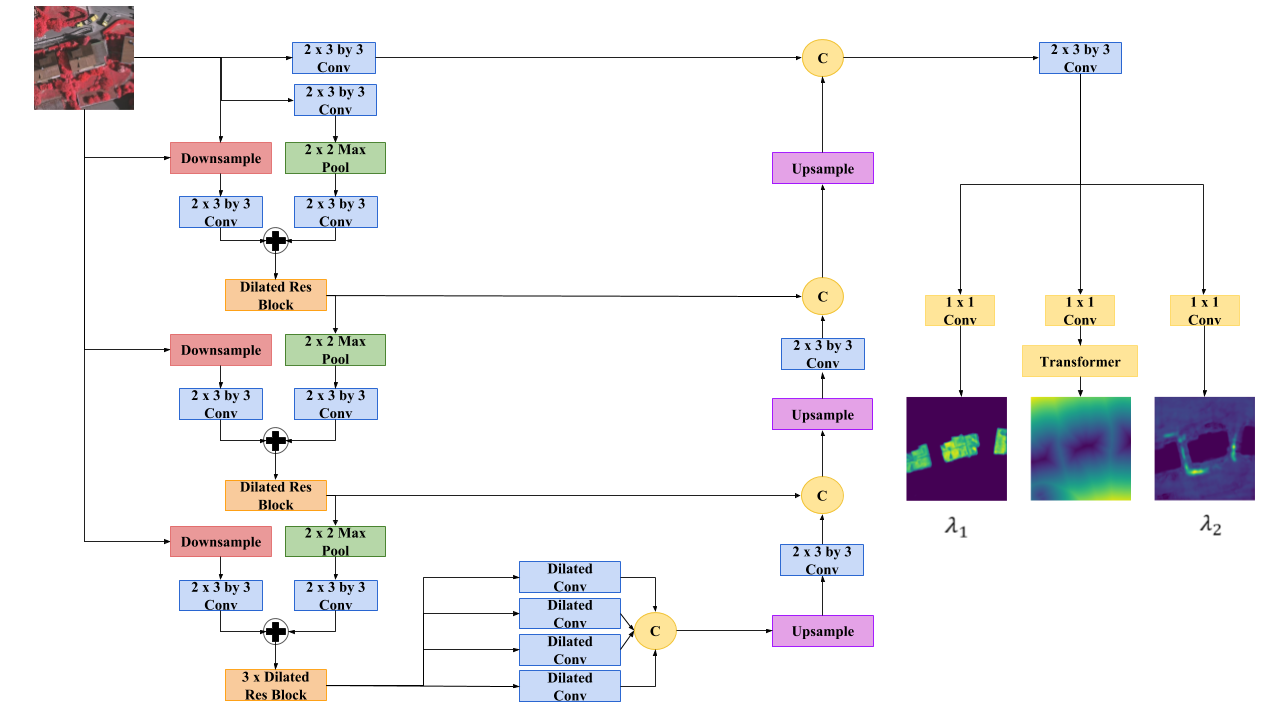}
\caption{Architecture of the CNN backbone.}
\label{fig:cnn}
\end{figure*}

As our CNN backbone, we follow \cite{hatamizadeh2019deepvessel} and utilize a fully convolutional encoder-decoder architecture with dilated residual blocks (Fig.~\ref{fig:cnn}). Each
convolutional layer is followed by a Rectified Linear Unit (ReLU) as
the activation layer and a batch normalization. The dilated residual
block consists of 2 consecutive dilated convolutional layers whose
outputs are fused with its input and fed into the ReLU activation
layer. In the encoder, each path consist of 2 consecutive $3 \times 3$
convolutional layers, followed by a dilated residual unit with a
dilation rate of 2. Before being fed into the dilated residual unit,
the output of these convolutional layers are added with the output
feature maps of another 2 consecutive $3 \times 3$ convolutional
layers that learn additional multi-scale information from the resized
input image in that resolution. To recover the content lost in the
learned feature maps during the encoding process, we utilize a series
of consecutive dilated residual blocks with dilation rates of 1, 2,
and 4 and feed the output to a dilated spatial pyramid pooling layer
with 4 different dilation rates of 1, 6, 12 and 18. The decoder is
connected to the dilated residual units at each resolution via skip
connections, and in each path we up-sample the image and employ 2
consecutive $3 \times 3$ convolutional layers before proceeding to the
next resolution. The output of the decoder is fed into another series
of 2 consecutive convolutional layer and then passed into 3 separate
$1 \times 1$ convolutional layers for predicting the output maps of
$\lambda_1$and $\lambda_2$ as well as the distance transform.


\section{DCAC Architecture and Implementation}

In our DCAC framework (Fig.~\ref{fig:framework}), the CNN backbone
serves to directly initialize the zero level-set contour as well as
the weighted local parameters. We initialize the zero level-set by a
learned distance transform that is directly predicted by the CNN along
with additional convolutional layers that learn the parameter maps.
Figure \ref{fig:leanred_func} illustrates an example of what the
backbone CNN learns in the DCAC on one input image from the Vaihingen
data set. These learned parameters are then passed to the ACM that
unfolds for a certain number of timesteps in a differentiable manner.
The final zero level-set is then converted to logits and compared with
the label and the resulting error is back-propagated through the
entire framework in order to tune the weights of the CNN backbone.
Algorithm~\ref{algorithm} presents the details of DCAC training
algorithm.

\begin{algorithm}[t]
 \caption{DCAC Training Algorithm}
 \KwData{$X$,$Y_\mathrm{gt}$: Paired image and label; $f$: CNN with parameters $\omega$; $g$: ACM with parameters $\lambda_1,\lambda_2$; $h$: Loss function; $N$: Number of ACM iterations; $\eta$: learning rate}
 \KwResult{$Y_\mathrm{out}$: Final segmentation}
 \While{not converged}{
  $\lambda_1,\lambda_2,\phi_{0}$=$f$($X$)\\
 \For{$t=1$ \textbf{to} $N$}{
    $\phi_{t}$=$g$($\phi_{t-1};\lambda_1,\lambda_2$)
    }
    $Y_{out}$=\hbox{Sigmoid}$(\phi_{N})$\\
  L=$h$$(Y_{out},Y_{gt})$\\
  compute $\frac{\partial L}{\partial \omega}$ and Back-propagate the error\\
  Update the Weights of $f$: $\omega \leftarrow \omega-\eta  \frac{\partial L}{\partial \omega}$
}
 \label{algorithm}
\end{algorithm}

\begin{figure*}
\subcaptionbox{Labeled
image}{\includegraphics[width=0.19\linewidth,height=0.19\linewidth]{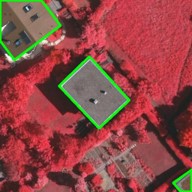}}\hfill
\subcaptionbox{DCAC, constant
$\lambda$s}{\includegraphics[width=0.19\linewidth,height=0.19\linewidth]{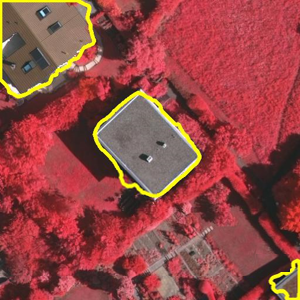}}\hfill
\subcaptionbox{DCAC}{\includegraphics[width=0.19\linewidth,height=0.19\linewidth]{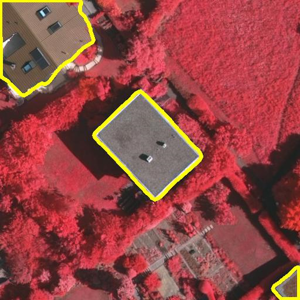}}\hfill
\subcaptionbox{$\lambda_1(x,y)$}{\includegraphics[width=0.19\linewidth,height=0.19\linewidth,trim={80
35 20 20},clip]{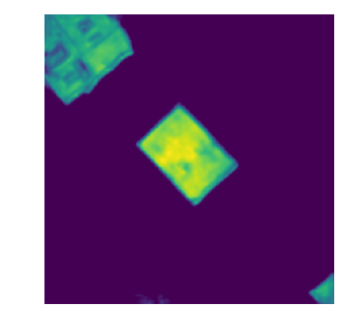}}\hfill
\subcaptionbox{$\lambda_2(x,y)$}{\includegraphics[width=0.19\linewidth,height=0.19\linewidth,trim={80
35 20 20},clip]{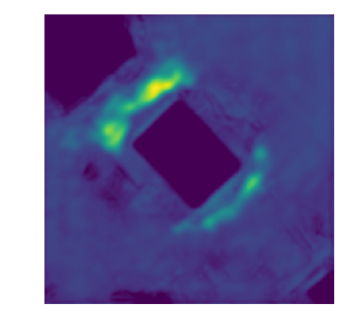}}
  \caption{(a) Labeled image (b) DCAC output with constant weighted
  parameters (c) DCAC output (d),(e) learned parameter maps
  $\lambda_{1}(x,y)$ and $\lambda_{2}(x,y)$ }
  \label{fig:const-lambda}
\end{figure*}

\subsection{Implementation Details}

All components of DCAC, including the ACM, have been implemented
entirely in Tensorflow \cite{abadi2016tensorflow} and are compatible
with both Tensorflow 1.x and 2.0 versions. The ACM implementation
benefits from the automatic differentiation utility of Tensorflow and
has been designed to enable the back-propagation of the error gradient
through the layers of the ACM.

In each ACM layer, each point along the the zero level-set contour is
probed by a local window and the mean intensity of the inside and
outside regions; i.e., $m_2$ and $m_1$ in (\ref{eq:f_pc}), are
extracted. In our implementation, $m_1$ and $m_2$ are extracted by
using a differentiable global average pooling layer with appropriate
padding not to lose any information on the edges.

All the training was performed on an Nvidia Titan XP GPU, and an
Intel® Core™ i7-7700K CPU @ 4.20GHz. The size of the minibatches for
training on the Vaihingen and Bing Huts datasets were 3 and 20
respectively. All the training sessions employ the Adam optimization
algorithm \cite{kingma2014adam} with a learning rate of 0.001 that
that decays by a factor of 10 every 10 epochs.

\begin{table*}[t]
\centering
\begin{tabular}{lcccc|cccc}
\toprule
\hfill Dataset:    & \multicolumn{4}{c}{Vaihingen} & \multicolumn{4}{|c}{Bing Huts}\\
    \midrule
Model    &   Dice  &   mIoU &   WCov  &  BoundF   &   Dice  &   mIoU &   WCov  &  BoundF  \\\midrule
DSAC     &    --       &     0.840   &	--    &	  --      &    --     &	  0.650  &	   --   &	--	\\[10pt]

UNet     &  0.810&	0.797&	0.843&	0.622& 0.710&	0.740&	0.852&	0.421  \\
ResNet  & 0.801	&0.791&	0.841&	0.770 & 0.81 &	0.797&	0.864&	0.434 \\
Backbone CNN    & 0.837	& 0.825 &	0.865 &	0.680 & 0.737 &	0.764 &	0.809 &	0.431  \\[10pt]

DCAC: Single Inst   &  \textbf{0.928} &	\textbf{0.929} &	\textbf{0.943} &	\textbf{0.819} & \textbf{0.855} &	\textbf{0.860}&	\textbf{0.894} &	\textbf{0.534} \\

DCAC: Multi Inst  &   \textbf{0.908} & \textbf{0.893} &	\textbf{0.910} &	\textbf{0.797} &  \textbf{0.797} &	\textbf{0.809}   & \textbf{0.839}    &	\textbf{0.491} \\[3pt]

DCAC: Single Inst, Const $\lambda$   &    0.877&	0.888	& 0.936 &	0.801      &     0.792 &	0.813	& 0.889&	0.513 \\

DCAC: Multi Inst, Const $\lambda$   &  0.857	&0.842	&0.876	&	0.707  &   0.757	& 0.777 & 0.891 & 0.486 \\
    \bottomrule
\end{tabular}
    \caption{Model Evaluations.}
    \label{table:datasets-perf}
\end{table*}

\section{Experiments}

\subsection{Datasets}

\paragraph{Vaihingen:}
The Vaihingen buildings
dataset \footnote{\url{http://www2.isprs.org/commissions/comm3/wg4/2d-sem-label-vaihingen.html}}
consists of 168 building images of size $512\times512$ pixels. The
labels for each image are generated by using a semi-automated
approach. We used 100 images for training and 68 for testing,
following the same data split as in \cite{marcos2018learning}. In this
dataset, almost all images consist of multiple instances of buildings,
some of which are located at the edges of the image.

\paragraph{Bing Huts:}
The Bing Huts
dataset \footnote{\url{https://www.openstreetmap.org/\#map=4/38.00/-95.80}}
consists of 605 images of size $64\times64$. We followed the same data
split that is used in \cite{marcos2018learning} and used 335 images
for training and 270 images for testing. This dataset is especially
challenging due the low spatial resolution and contrast that are
exhibited in the images.

\subsection{Evaluation Metrics and Loss Function}
To evaluate our model's performance, we utilized five different
metrics---Dice, mean Intersection over Union (mIoU), Weighted Coverage
(WCov), Boundary F (BoundF), and Root Mean Square Error (RMSE). The
original DSAC paper only reported on mIoU for both Vaihingen and Bing
Huts and only RMSE for the Bing Huts dataset. However, since the
delineation of boundaries is one of the important goals of our
framework, we employ the BoundF metric \cite{perazzi2016benchmark} to
precisely measure the similarity between the specific boundary pixels
in our predictions and the corresponding image labels. Furthermore, we
used a soft Dice loss function in training our model.

\section{Results and Discussion }

\begin{figure*}
\centering
\vskip -10pt \centerline{Vaihingen:\hfill}
\includegraphics[width=0.16\linewidth,height=0.16\linewidth]{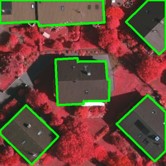}
\hfill
\includegraphics[width=0.16\linewidth,height=0.16\linewidth]{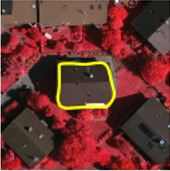}
\hfill
\includegraphics[width=0.16\linewidth,height=0.16\linewidth]{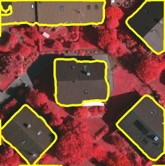}
\hfill
\includegraphics[width=0.16\linewidth,height=0.16\linewidth,trim={80
35 20 20},clip]{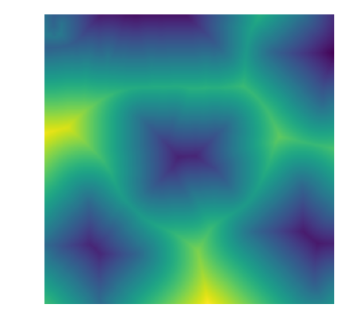} \hfill
\includegraphics[width=0.16\linewidth,height=0.16\linewidth,trim={80
35 20 20},clip]{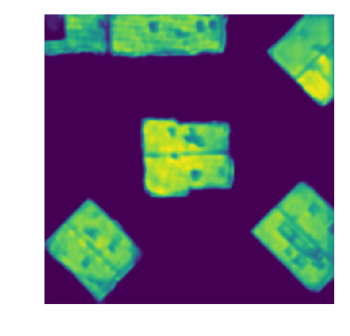} \hfill
\includegraphics[width=0.16\linewidth,height=0.16\linewidth,trim={80
35 20 20},clip]{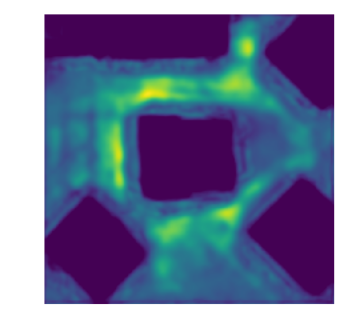}\\[4pt]

\includegraphics[width=0.16\linewidth,height=0.16\linewidth]{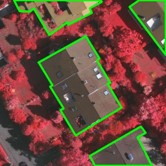}
\hfill
\includegraphics[width=0.16\linewidth,height=0.16\linewidth]{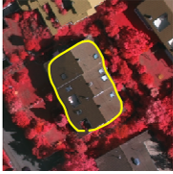}
\hfill
\includegraphics[width=0.16\linewidth,height=0.16\linewidth]{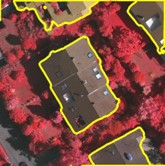}
\hfill
\includegraphics[width=0.16\linewidth,height=0.16\linewidth,trim={80
35 20 20},clip]{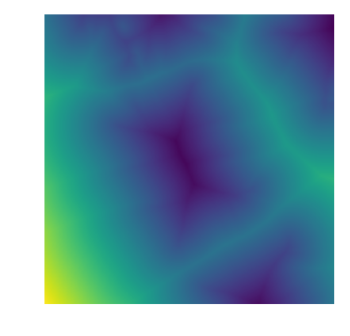} \hfill
\includegraphics[width=0.16\linewidth,height=0.16\linewidth,trim={80
35 20 20},clip]{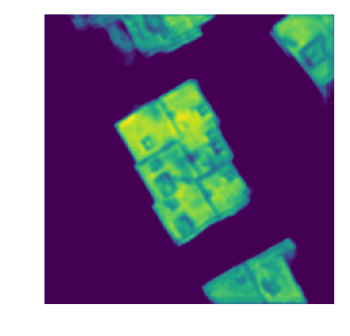} \hfill
\includegraphics[width=0.16\linewidth,height=0.16\linewidth,trim={80
35 20 20},clip]{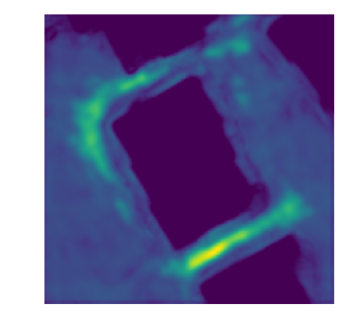}\\[4pt]

\includegraphics[width=0.16\linewidth,height=0.16\linewidth]{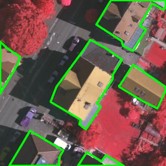}
\hfill
\includegraphics[width=0.16\linewidth,height=0.16\linewidth]{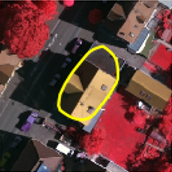}
\hfill
\includegraphics[width=0.16\linewidth,height=0.16\linewidth]{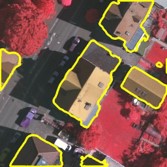}
\hfill
\includegraphics[width=0.16\linewidth,height=0.16\linewidth,trim={80
35 20 20},clip]{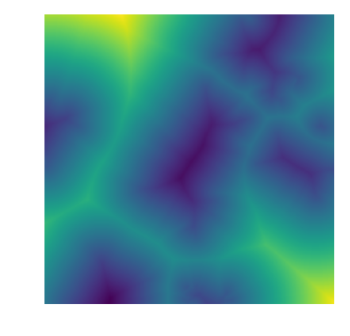} \hfill
\includegraphics[width=0.16\linewidth,height=0.16\linewidth,trim={80
35 20 20},clip]{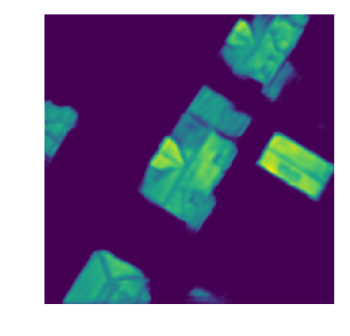} \hfill
\includegraphics[width=0.16\linewidth,height=0.16\linewidth,trim={80
35 20 20},clip]{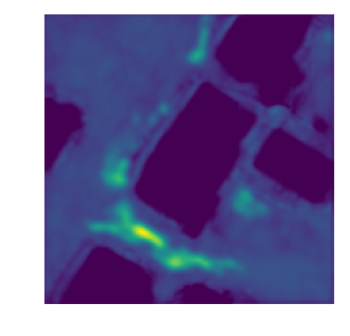}\\[4pt]

\includegraphics[width=0.16\linewidth,height=0.16\linewidth]{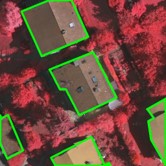}
\hfill
\includegraphics[width=0.16\linewidth,height=0.16\linewidth]{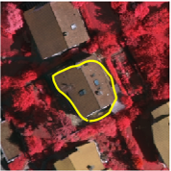}
\hfill
\includegraphics[width=0.16\linewidth,height=0.16\linewidth]{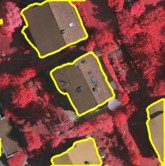}
\hfill
\includegraphics[width=0.16\linewidth,height=0.16\linewidth,trim={80
35 20 20},clip]{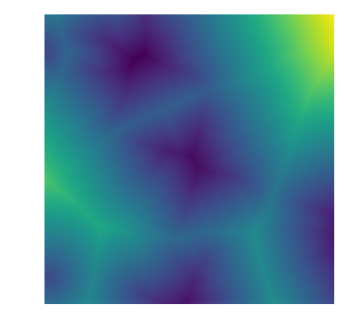} \hfill
\includegraphics[width=0.16\linewidth,height=0.16\linewidth,trim={80
35 20 20},clip]{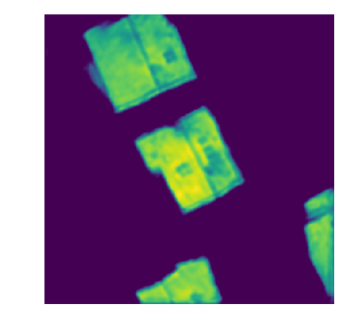} \hfill
\includegraphics[width=0.16\linewidth,height=0.16\linewidth,trim={80
35 20 20},clip]{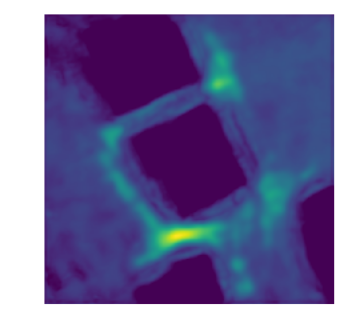}\\[4pt]

\centerline{Bing Huts:\hfill}
\includegraphics[width=0.16\linewidth,height=0.16\linewidth]{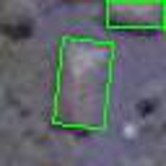}
\hfill
\includegraphics[width=0.16\linewidth,height=0.16\linewidth]{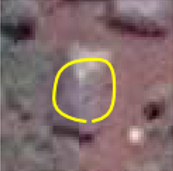}
\hfill
\includegraphics[width=0.16\linewidth,height=0.16\linewidth]{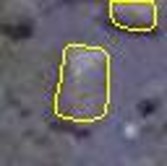}
\hfill
\includegraphics[width=0.16\linewidth,height=0.16\linewidth]{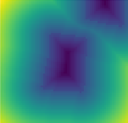}
\hfill
\includegraphics[width=0.16\linewidth,height=0.16\linewidth]{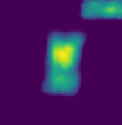}
\hfill
\includegraphics[width=0.16\linewidth,height=0.16\linewidth]{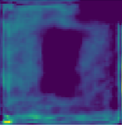}\\[4pt]

\includegraphics[width=0.16\linewidth,height=0.16\linewidth]{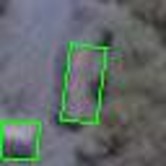}
\hfill
\includegraphics[width=0.16\linewidth,height=0.16\linewidth]{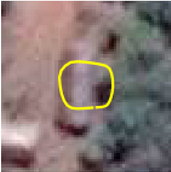}
\hfill
\includegraphics[width=0.16\linewidth,height=0.16\linewidth]{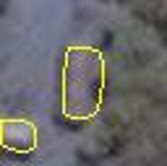}
\hfill
\includegraphics[width=0.16\linewidth,height=0.16\linewidth]{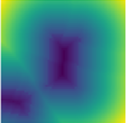}
\hfill
\includegraphics[width=0.16\linewidth,height=0.16\linewidth]{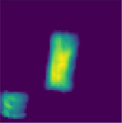}
\hfill
\includegraphics[width=0.16\linewidth,height=0.16\linewidth]{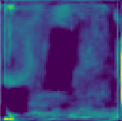}\\[4pt]

\includegraphics[width=0.16\linewidth,height=0.16\linewidth]{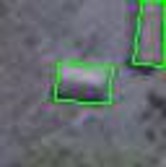}
\hfill
\includegraphics[width=0.16\linewidth,height=0.16\linewidth]{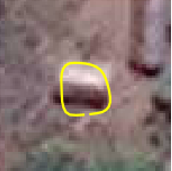}
\hfill
\includegraphics[width=0.16\linewidth,height=0.16\linewidth]{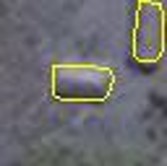}
\hfill
\includegraphics[width=0.16\linewidth,height=0.16\linewidth]{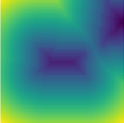}
\hfill
\includegraphics[width=0.16\linewidth,height=0.16\linewidth]{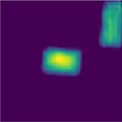}
\hfill
\includegraphics[width=0.16\linewidth,height=0.16\linewidth]{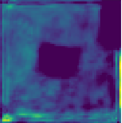}\\[4pt]

\makebox[0.16\textwidth]{\hskip -5mm (a) Labeled Image}
\makebox[0.16\textwidth]{(b) DSAC} \makebox[0.16\textwidth]{(c) Our
DCAC} \makebox[0.16\textwidth]{(d) DT} \makebox[0.16\textwidth]{(e)
$\lambda_1(x,y)$} \makebox[0.16\textwidth]{(f) $\lambda_2(x,y)$}
\caption{Comparative visualization of the labeled image, the output of
DSAC, and the output of our DCAC, for the Vaihingen (top) and Bing
Huts (bottom) datasets: (a) Image with label (green), (b) DSAC output,
(c) our DCAC output, (d) DCAC learned distance transform, (e)
$\lambda_{1}$ and (f) $\lambda_{2}$ for the DCAC.}
  \label{fig:final_comparison}
\end{figure*}

\subsection{Local and Fixed Weighted Parameters}

To validate the contribution of the local weighted parameters in the
level-set ACM, we also trained our DCAC on both the Vaihingen and Bing
Huts datasets by only allowing one trainable scalar parameter for each
of $\lambda_1$ and $\lambda_2$, which is constant over the entire
image. As presented in Table~\ref{table:datasets-perf}, in both the
Vaihingen and Bing Huts datasets, this constant-$\lambda$ formulation
still outperforms the baseline CNN in all evaluation metrics for both
single-instance and multi-instance buildings, thus showing the
effectiveness of the end-to-end training of the DCAC. However, the
DCAC with the full $\lambda_1(x,y)$ and $\lambda_2(x,y)$ maps
outperforms this constant formulation by a wide margin in all
experiments and metrics.

A key metric of interest in this comparison is the BoundF score, which
demonstrates how our local formulation captures the details of the
boundaries more effectively by adjusting the inward and outward forces
on the contour locally. As illustrated in Figure
\ref{fig:const-lambda}, DCAC has perfectly delineated the boundaries
of the building instances. However, DCAC with constant formulation has
over-segmented these instances.


\subsection{Buildings on the Edges of the Image}

Our DCAC is capable of properly segmenting the instances of buildings
located on the edges of some of the images present in the Vaihingen
dataset. This is mainly due to the proper padding scheme that we have
utilized in our global average pooling layer used to extract the local
intensities of pixels while avoiding the loss of information on the
boundaries.

\subsection{Initialization and Number of ACM Iterations}

In all cases, we performed our experiments with the goal of leveraging
the CNN to fully automate the ACM and eliminate the need for any human
supervision. Our scheme for learning a generalized distance transform
directly helped us to localize all the building instances
simultaneously and initialize the zero level-sets appropriately while
avoiding a computationally expensive and non-differentiable distance
transform operation. In addition, initializing the zero level-sets in
this manner, instead of the common practice of initializing from a
circle, helped the contour to converge significantly faster and avoid
undesirable local minima.

\subsection{Comparison Against the DSAC Model}

Although most of the images in the Vaihingen dataset consist of
multiple instances of buildings, the DSAC model
\cite{marcos2018learning} can deal with only a single building at a
time. For a fair comparison between the two approaches, we report
separate metrics for a single building, as reported by in
\cite{marcos2018learning} for the DSAC, as well as for all the
instances of buildings (which the DSAC cannot handle). As presented in
Table~\ref{table:datasets-perf}, our DCAC outperforms DSAC by $7.5$
and $21$ percent in mIoU respectively on both the Vaihingen and Bing
Huts datasets. Furthermore, the multiple-instance metrics of our DCAC
outperform the single-instance DSAC results. As demonstrated in
Fig.~\ref{fig:final_comparison}, in the Vaihingen dataset, DSAC
struggles in coping with the topological changes of the buildings and
fails to appropriately capture sharp edges, while our framework in
most cases handles these challenges. In the Bing Hut dataset, the DSAC
is able to localize the buildings, but it mainly over-segments the
buildings in many cases. This may be due to DSAC's inability to
distinguish the building from the surrounding soil because of the low
contrast and small size of the image. By comparison, our DCAC is able
to low contrast dataset well, with more accurate boundaries, when
comparing the segmentation output of DSAC (b) and our DCAC (c), as
seen in Fig.~\ref{fig:final_comparison}.

\section{Conclusions and Future Work}

We have introduced a novel image segmentation framework, called DCAC,
which is a truly end-to-end integration of ACMs and CNNs. We proposed
a novel locally-penalized Eulerian energy model that allows for
pixel-wise learnable parameters that can adjust the contour to
precisely capture and delineate the boundaries of objects of interest
in the image. We have tackled the problem of building instance
segmentation on two very challenging datasets of Vaihingen and Bing
Huts as test case and our model outperforms the current
state-of-the-art method, DSAC. Unlike DSAC, which relies on the manual
initialization of its ACM contour, our model requires minimal human
supervision and is initialized and guided by its CNN backbone.
Moreover, DSAC can only segment a single building at a time whereas
our DCAC can segment multiple buildings simultaneously. We also showed
that, unlike DSAC, our DCAC is effective in handling various
topological changes in the image. Given the level of success that DCAC
has achieved in this application and the fact that it features a
general Eulerian formulation, it is readily applicable to other
segmentation tasks in various domains where purely CNN filter-based
approaches can benefit from the versatility and precision of ACMs in
delineating object boundaries in images.

{\small
\bibliographystyle{ieee}
\bibliography{iccv19}
}

\end{document}